\documentclass[sn-nature]{sn-jnl}

\usepackage{graphicx}%
\usepackage{multirow}%
\usepackage{amsmath,amssymb,amsfonts}%
\usepackage{amsthm}%
\usepackage{mathrsfs}%
\usepackage[title]{appendix}%
\usepackage{xcolor}%
\usepackage{textcomp}%
\usepackage{manyfoot}%
\usepackage{booktabs}%
\usepackage{algorithm}%
\usepackage{algorithmicx}%
\usepackage{algpseudocode}%
\usepackage{textcomp}
\usepackage{listings}%
\usepackage{url}  
\usepackage{hyperref}
\usepackage[version=4]{mhchem} 

\begin{document}

\title[Enzymatic Reaction Recommendation with Active Learning]{RXNRECer Enables Fine-grained Enzymatic Function Annotation through Active Learning and Protein Language Models}


\author[1]{\fnm{Zhenkun} \sur{Shi}}
\author[2]{\fnm{Jun} \sur{Zhu}}
\author[1]{\fnm{Dehang} \sur{Wang}}
\author[2]{\fnm{BoYu} \sur{Chen}}
\author[1]{\fnm{Qianqian} \sur{Yuan}}
\author[1]{\fnm{Zhitao} \sur{Mao}}
\author[1]{\fnm{Fan} \sur{Wei}}
\author*[2]{\fnm{Weining} \sur{Wu}}\email{wuweining@hrbeu.edu.cn}
\author*[1]{\fnm{Xiaoping} \sur{Liao}}\email{liao\_xp@tib.cas.cn}
\author*[1]{\fnm{Hongwu} \sur{Ma}}\email{ma\_hw@tib.cas.cn}

\affil*[1]{\orgdiv{Biodesign Center}, \orgname{Key Laboratory of Engineering Biology for Low-carbon Manufacturing, Tianjin Institute of Industrial Biotechnology, Chinese Academy of Sciences}, \orgaddress{\postcode{300308}, \state{Tianjin}, \country{China}}}

\affil[2]{\orgdiv{College of Computer Science and Technology}, \orgname{Harbin Engineering University}, \orgaddress{\street{Street}, \city{Harbin}, \postcode{10587}, \state{State}, \country{China}}}

\abstract{

A key challenge in enzyme annotation is identifying the biochemical reactions catalyzed by proteins. Most existing methods rely on Enzyme Commission (EC) numbers as intermediaries, which first predict an EC number and then retrieve associated reactions. This indirect strategy introduces ambiguity due to the complex many-to-many mappings among proteins, EC numbers, and reactions, and is further complicated by frequent updates to EC numbers and inconsistencies across databases. To address these challenges, we present \textit{RXNRECer}, a transformer-based ensemble framework that directly predicts enzyme-catalyzed reactions without relying on EC numbers. It integrates protein language modeling and active learning to capture both high-level sequence semantics and fine-grained transformation patterns. Evaluations on curated cross-validation and temporal test sets demonstrate consistent improvements over six EC-based baselines, with gains of 16.54\% in F1 score and 15.43\% in accuracy. Beyond accuracy gains, the framework offers clear advantages for downstream applications, including scalable proteome-wide reaction annotation, enhanced specificity in refining generic reaction schemas, systematic annotation of previously uncurated proteins, and reliable identification of enzyme promiscuity. By incorporating large language models, it also provides interpretable prediction rationales. These capabilities make \textit{RXNRECer} a robust and versatile solution for EC-free, fine-grained enzyme function prediction, with potential applications across multiple areas of enzyme research and industrial applications.
}


\maketitle

\section{Introduction}\label{sec1}

Proteins play essential roles in cellular activities, underpinning biological growth, development, genetic transmission, and the execution of vital functions. Systematic annotation of protein functions is critical for understanding life processes at the molecular level and significantly advances research in protein interaction mechanisms\cite{burke2019design}, biosynthetic pathway design\cite{chen2020data, watanabe2009escherichia}, metabolic engineering\cite{papp2004metabolic, hatzimanikatis2004metabolic, mao2022ecmpy}, and biopharmaceutical development\cite{shu2019reaction}. Among protein annotation tasks, identifying enzyme–reaction relationships is particularly crucial, as enzymes catalyze thousands of chemical reactions fundamental to biological systems.

Recent advancements in sequencing technologies have dramatically accelerated the accumulation of protein sequence data. However, the traditional reliance on experimental methods for protein function annotation has widened the gap between available sequence data and accurately annotated protein functions. For instance, in June 2025 alone, UniProt added 2,299,276 protein sequences to TrEMBL, whereas only 434 sequences were manually curated and incorporated into the SwissProt database. This disparity underscores the urgent need for scalable computational methods. Consequently, computational approaches that utilize existing protein–reaction knowledge to predict functions of uncharacterized proteins have gained prominence.

The Enzyme Commission (EC) number system, developed by the IUBMB, classifies enzymes based on the chemical reactions they catalyze. Each EC number represents a specific enzymatic function and provides a hierarchical framework that links enzymes to broad reaction categories. Owing to its structured format and ease of access, the EC system has become a widely adopted standard for enzyme function annotation. Many existing methods \cite{ryu2019deep, claudel2003enzyme, yu2009genome, dalkiran2018ecpred, shi2022ecrecer, yu2023enzyme} formulate this task as a classification problem using EC numbers as target labels, as illustrated in Supplementary Fig.~S4(a). However, while EC numbers provide a structured way to annotate enzyme functions, they primarily encode reaction classes rather than specific biochemical reactions. Consequently, mapping an enzyme to a precise reaction equation often requires additional manual alignment with biochemical databases, a labor-intensive process that is prone to errors.

Further complicating matters, EC classifications are frequently updated, involving reassignment (e.g., EC 1.1.1.68 $\rightarrow$ EC 1.5.1.20), splitting (e.g., EC 1.1.1.246 $\rightarrow$ EC 1.1.1.348 and EC 4.2.1.139), and deletion (e.g., EC 2.1.1.30). Moreover, inconsistencies arise due to asynchronous update frequencies across different databases. For instance, EC 4.2.1.13, created in 1961 and reassigned to EC 4.3.1.17 in 2001, still had 186 protein entries annotated under the obsolete EC in SwissProt as of March 2022, linked to the reaction \ce{$\beta$-D-fructose 1,6-bisphosphate -> D-glyceraldehyde 3-phosphate + dihydroxyacetone phosphate}. In contrast, the updated EC 4.3.1.17, associated with the reaction \ce{L-serine -> NH4+ + pyruvate}, had only 49 annotated entries. According to EC classification standards, these annotations should have been unified; however, due to lagging database updates, outdated annotations persist, leading to inconsistencies in enzyme–reaction relationships.

To address these challenges, there is a growing need for methods that can directly predict the specific biochemical reactions catalyzed by enzymes, rather than relying on EC numbers as intermediaries (Supplementary Fig.~S4(a)). Direct reaction-level prediction (Supplementary Fig.~S4(b)) bypasses the ambiguity and update inconsistencies inherent in EC-based systems, while also eliminating the labor-intensive step of manually linking enzymes to reactions via external databases. By shifting the predictive target from coarse-grained EC classes to fine-grained reaction representations, such approaches provide a more accurate and scalable solution for enzyme function annotation, while remaining compatible with existing EC-based knowledge when needed.

In this work, we propose \textit{RXNRECer}, a transformer-based dynamic ensemble learning framework for direct enzyme–reaction prediction. It integrates a protein language model (PLM) to capture high-level sequence semantics, an active learning strategy for efficient and targeted fine-tuning, and a dynamic ensemble module that emphasizes direct reaction-level prediction while selectively incorporating EC-based and PLM similarity-based signals to improve robustness across diverse reaction types. The framework also features an interpretability component powered by a general-purpose language model, enabling transparent and biologically meaningful insights into predicted functions. Extensive evaluations across curated cross-validation sets and independent temporal test data demonstrate that the method consistently outperforms traditional MSA-based approaches, EC-based tools, and recent PLM-based baselines, providing a scalable and generalizable solution for fine-grained enzyme function annotation.

Beyond benchmark evaluations, we further applied RXNRECer to diverse real-world scenarios, including full-proteome annotation, substrate-specific function resolution, uncharacterized protein analysis, and promiscuous enzyme prediction. These applications demonstrate its versatility across practical biological contexts.

\section{Results}\label{sec2}

\begin{figure}[htbp!]
  \centering	
  \includegraphics[width=1.0\textwidth]{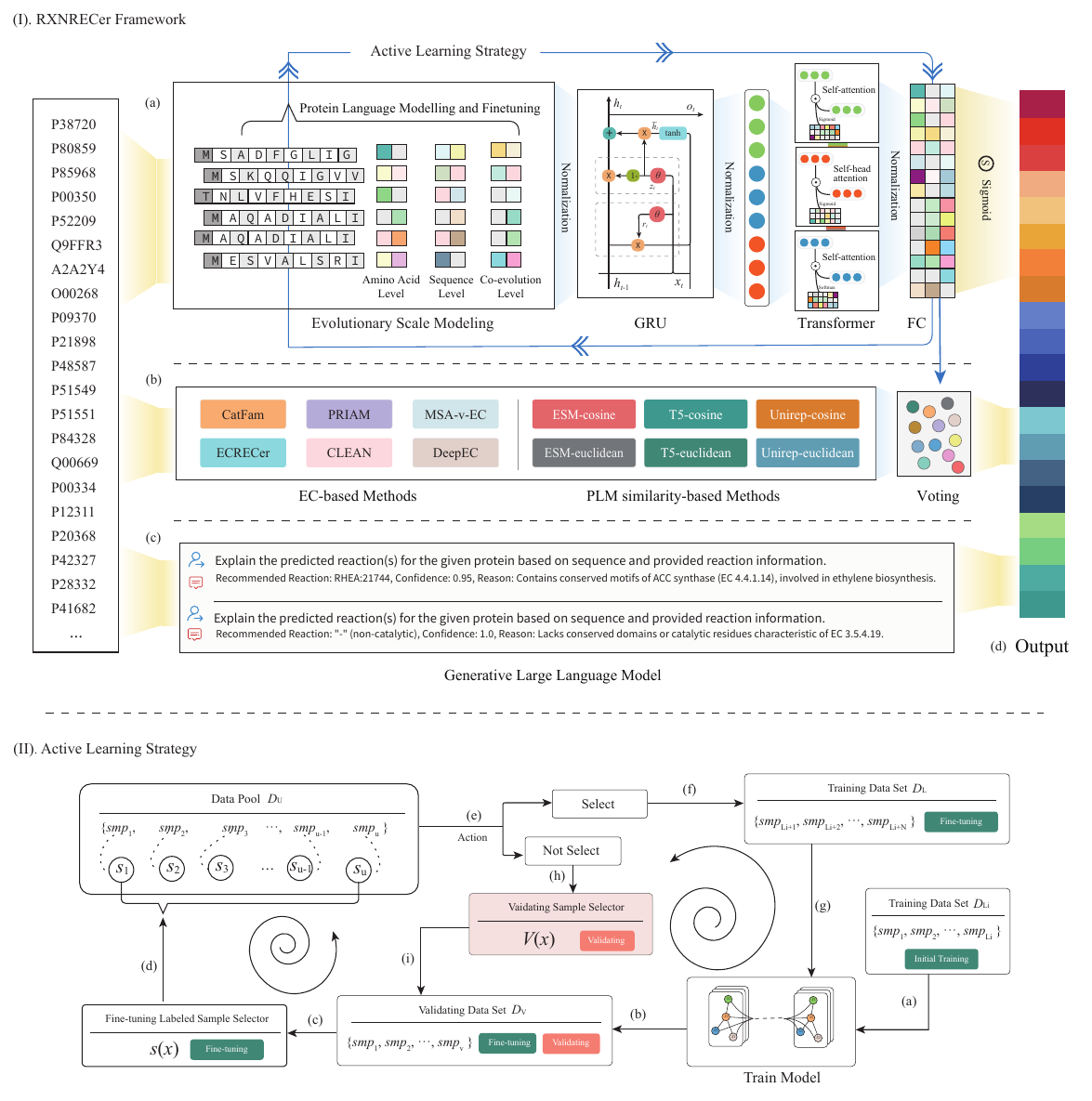}
  \caption{
    Schematic overview of the RXNRECer framework and its active learning workflow.
    (I) \textbf{RXNRECer Framework.} 
    (a) \textit{Sequence-based reaction prediction (RXNRECer-S1).} Protein sequences are embedded using a pre-trained language model. The resulting multi-level embeddings are passed through GRU and Transformer layers, followed by a fully connected (FC) layer to produce reaction scores. This EC-independent prediction module is fine-tuned via a reaction-level classification task within an active learning framework. 
    (b) \textit{Ensemble integration (RXNRECer-S2).} Predictions from EC-based and PLM similarity-based methods are combined with the direct predictor using a voting scheme to improve robustness across diverse reaction types. 
    (c) \textit{Interpretation and ranking (RXNRECer-S3).} A general-purpose language model (GLM) is prompted to re-rank the candidate reactions and generate concise, biologically meaningful justifications.
    (II) \textbf{Active Learning Strategy.} 
    (a) \textit{Initial model training.} The model is initially trained on a small labeled set $D_{Li}$. 
    (b) \textit{Validation.} Performance is evaluated on a validation set $D_V$ sampled from the unlabeled pool $D_U$. 
    (c–e) \textit{Sample selection.} An acquisition function $S(x)$ identifies the most informative samples for labeling. 
    (f–g) \textit{Fine-tuning.} The selected samples are incorporated into the training set $D_L$ to refine the model. 
    (h–i) \textit{Validation sampling.} Additional samples are selected to monitor performance after each fine-tuning iteration.
  }
  \label{fig:framework}
\end{figure}

We present a comprehensive evaluation of the proposed framework, covering both systematic performance assessments and diverse application scenarios. We begin with a detailed overview of RXNRECer, including its architectural components, ensemble integration, and active learning workflow (Fig.~\ref{fig:framework}). We then benchmark its predictive accuracy against EC-based and PLM-based baselines under two settings: 10-fold cross-validation and independent temporal testing. Further analyses explore the impact of dynamic ensemble strategies and active learning on performance gains. To enhance interpretability, we incorporate structured prompts and a general-purpose language model (GLM) to provide biologically meaningful rationales. Finally, we demonstrate the versatility of the framework through four case studies, including proteome-scale annotation in \textit{Fusarium venenatum}, reaction schema refinement, annotation of proteins without curated catalytic information, and uncovering of enzyme promiscuity.

\subsection{Overview of RXNRECer}

The RXNRECer framework comprises three key components: a PLM-based reaction classifier (RXNRECer-S1), a dynamic ensemble prediction module (RXNRECer-S2), and a GLM-based interpretability module (RXNRECer-S3) (Fig.~\ref{fig:framework}).

The first component (RXNRECer-S1; Fig.~\ref{fig:framework} I(a)) is a PLM-based reaction classifier that directly predicts the biochemical reactions catalyzed by a given enzyme. Protein sequences are embedded using the ESM2-650M model to capture amino acid-level, sequence-level, and co-evolutionary information. These embeddings are processed through a Gated Recurrent Unit (GRU) layer to model long-range dependencies, followed by a Transformer layer to capture short-range contextual features. A fully connected (FC) layer with a sigmoid activation function generates reaction prediction scores. The entire architecture, including the embedding layers, is jointly fine-tuned through a multi-label classification task under an active learning framework (Fig.~\ref{fig:framework} II).

The second component (RXNRECer-S2; Fig.\ref{fig:framework} I(b)) is a dynamic ensemble prediction module that integrates predictions from heterogeneous sources to enhance robustness. It incorporates outputs from multiple EC-based methods (e.g., CatFam\cite{yu2009genome}, ECRECer~\cite{shi2022ecrecer}, PRIAM~\cite{claudel2003enzyme}, DeepEC~\cite{ryu2019deep}, CLEAN~\cite{yu2023enzyme}) and PLM similarity-based methods (e.g., T5-cosine~\cite{prostT5}, Unirep-cosine~\cite{unirep}, ESM-cosine~\cite{lin2023evolutionary}). These predictions are dynamically aggregated using three complementary ensemble strategies: (1) majority voting, which selects reactions with the highest agreement across methods; (2) stacking, which trains a meta-classifier to combine top-performing models for improved accuracy; and (3) recall-boost, which unifies all candidate reactions to maximize coverage in recall-sensitive applications. This flexible integration mechanism mitigates the limitations of individual models and adapts to different prediction scenarios.

The final component (RXNRECer-S3; Fig.~\ref{fig:framework} I(c)) is a GLM-based interpretability module that contextualizes the candidate reactions. It takes ensemble outputs along with enzyme-side inputs (e.g., protein sequence or available database identifiers) and reaction-side information (e.g., reaction equations, structured identifiers, or molecular representations) as input. Guided by a structured prompt, a GLM prioritizes reactions by likelihood and generates concise, biologically grounded explanations. This dual function enables both transparent interpretation and mechanistic insight into enzyme–reaction relationships. This integrated framework combines prediction accuracy, flexibility, and interpretability, laying a solid foundation for the following empirical evaluations.

\subsection{Cross-Validation Evaluation: 10-Fold Assessment of RXNRECer and Baselines}

\begin{figure}[htbp!]
  \centering	
  \includegraphics[width=1.0\textwidth]{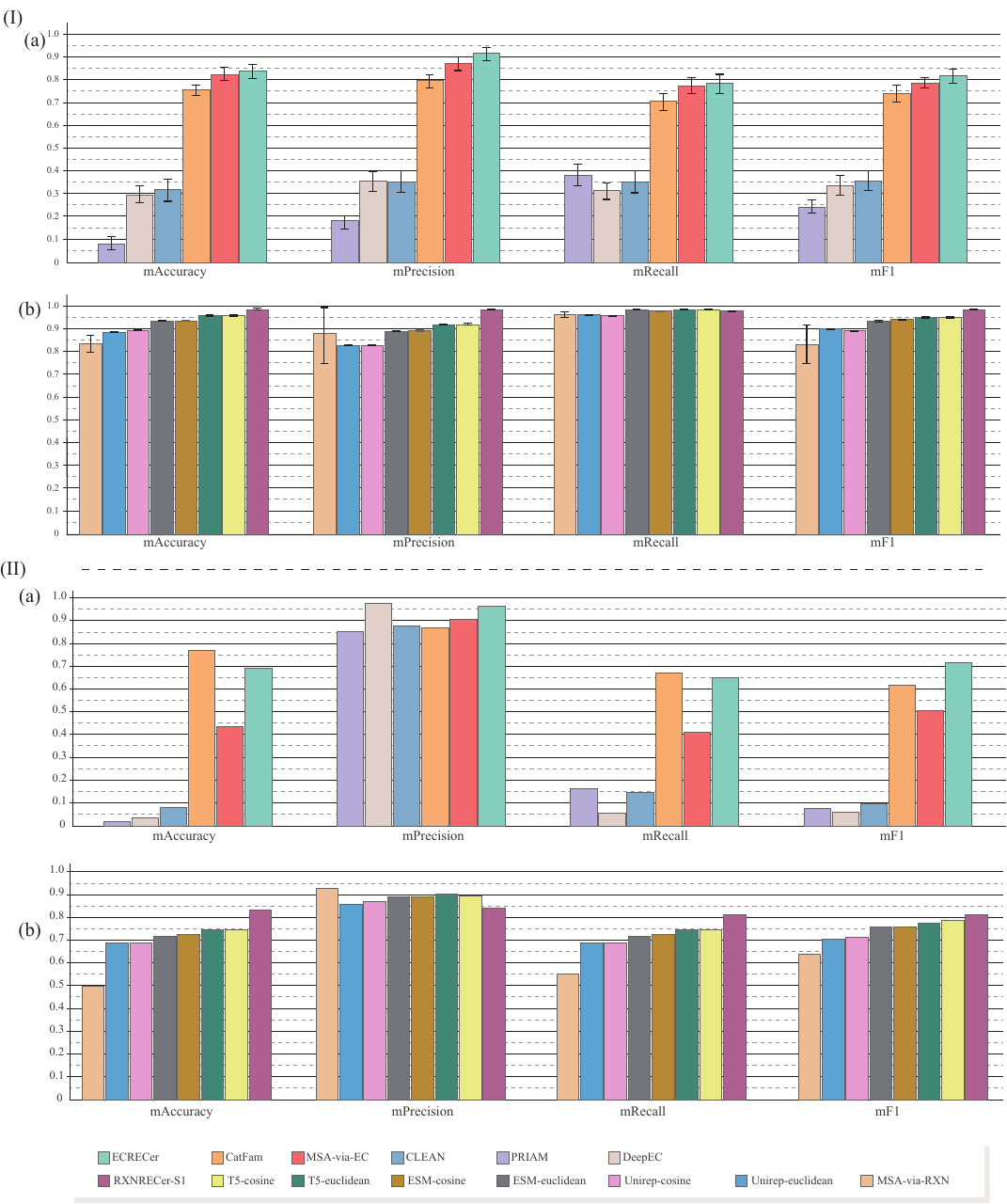}
  \caption{
    Comparative evaluation of enzyme reaction prediction methods on two datasets.
    (I) Results on the 10-fold cross-validation dataset (\textit{ds\_rcv}). 
    (II) Results on the independent temporal test dataset (\textit{ds\_rcp}).
    For each panel: 
    (a) Performance of EC-based methods (ECRECer, CatFam, CLEAN, PRIAM, DeepEC, and MSA-via-EC).
    (b) Performance of PLM-based similarity methods (T5-cosine, T5-euclidean, ESM-cosine, ESM-euclidean, Unirep-cosine, Unirep-euclidean), the end-to-end fine-tuned RXNRECer-S1, and MSA-via-RXN.
	Results on \textit{ds\_rcv} are averaged over 10 folds with error bars showing standard deviations; results on \textit{ds\_rcp} are from a single evaluation without error bars.
  }
  \label{fig:res_10_folds}
\end{figure}

We evaluated various enzyme reaction prediction strategies using a 10-fold cross-validation dataset (\textit{ds\_rcv}), covering EC-based annotation methods, unsupervised similarity-based models, and direct reaction-level predictors. As shown in Fig.~\ref{fig:res_10_folds}(I) and detailed in Supplementary Tables S3–S5, the benchmark includes six EC-based methods (ECRECer, CLEAN, DeepEC, PRIAM, CatFam, MSA-via-EC), six PLM-based similarity approaches (T5, ESM, UniRep with cosine or Euclidean distance), a sequence alignment baseline (MSA-via-RXN), and our proposed model, RXNRECer-S1.

Among all EC-based methods evaluated (Fig.~\ref{fig:res_10_folds}(I)(a)), ECRECer achieved the best performance, with an mAccuracy of 83.30\% ± 2.71\% and an mF1 of 81.93\% ± 2.65\%, benefiting from a systematically constructed annotation framework. MSA-via-EC (78.28\% ± 2.71\%) and CatFam (74.06\% ± 3.44\%) followed, the latter aided by profile calibration and controlled false-positive rates. In contrast, CLEAN, DeepEC, and PRIAM showed substantially lower performance (all metrics below 40\%), with DeepEC and PRIAM further limited by incomplete coverage—failing to annotate over 30\% of test proteins on average (see Supplementary Table S3). Despite ECRECer’s relative strength, it was still outperformed by the direct sequence alignment method MSA-via-RXN (mF1: 82.92\% ± 8.09\%), which bypasses EC numbers and directly links proteins to reactions. Notably, MSA-via-RXN also exceeded its EC-based counterpart MSA-via-EC, reinforcing the advantage of reaction-level prediction. These results highlight a core limitation of EC-based strategies: the reliance on hierarchical EC annotations as intermediaries introduces abstraction and coverage gaps, which reduce their effectiveness for comprehensive and precise reaction inference.

To explore alternatives to EC-based annotation, we assessed unsupervised similarity-based strategies that combine protein language model (PLM) embeddings with either cosine or Euclidean distance. These methods rely solely on pretrained representations to estimate functional similarity in embedding space, without using labeled reactions or EC numbers. As shown in Fig.~\ref{fig:res_10_folds}(I)(b), across all PLMs examined (UniRep, ESM, and T5), cosine similarity consistently resulted in slightly better performance than Euclidean distance, potentially due to its sensitivity to angular relationships that better reflect conformational or functional similarities in proteins \cite{draganov2024hidden, pillai2024novo}. For instance, cosine-based variants achieved marginally higher scores in both mF1 and mRecall across all models: ESM-cosine reached an mF1 of 93.52\% and mRecall of 98.03\%, compared to 93.47\% and 97.99\% with Euclidean; UniRep-cosine achieved an mF1 of 88.98\%, slightly above its Euclidean counterpart (88.87\%). These methods also provided complete coverage (no missing predictions) and low standard deviation across folds (e.g., ESM-cosine mAccuracy std: 0.0014), indicating stable and generalizable performance (see Supplementary Table S4). Compared to the best EC-based method (ECRECer), these unsupervised PLM similarity models improved mAccuracy by 6.94\%–13.00\% and mF1 by 7.05\%–13.00\%, and also exceeded the reaction-level alignment baseline MSA-via-RXN by 5.95\%–12.01\% in mF1. These results demonstrate that PLMs encode informative representations for enzyme functions and support high-quality predictions even in the absence of supervised training.

Building on these results, we further fine-tuned the ESM model using curated protein–reaction pairs, integrating additional GRU and Transformer layers to build RXNRECer Stage 1 (RXNRECer-S1). Unlike unsupervised approaches, RXNRECer-S1 learns to directly associate protein sequences with reaction labels through end-to-end optimization. Among all configurations tested, the ESM-based RXNRECer-S1 achieved the best overall performance, with an mAccuracy of 98.73\% ± 0.06\%, mPrecision of 98.90\% ± 0.09\%, mRecall of 98.04\% ± 0.13\%, and mF1 of 98.47\% ± 0.10\% (see Supplementary Table S5). These results represent a substantial improvement over both the best EC-based method (ECRECer, mF1: 81.93\%) and the top unsupervised similarity model (e.g., ESM-cosine, mF1: 93.52\%), corresponding to gains of +16.5 and +4.95 percentage points in mF1, respectively. RXNRECer-S1 also outperformed the alignment-based method MSA-via-RXN (mF1: 82.92\%) by a wide margin, while maintaining full coverage and low variance across cross-validation folds.

Overall, on the 10-fold cross-validation dataset, RXNRECer-S1 consistently delivered the highest accuracy, recall, and F1 scores among all strategies evaluated, demonstrating the advantage of directly modeling enzyme–reaction relationships through supervised fine-tuning of pretrained protein language models.

\subsection{Real-World Generalization Evaluation: Performance on Recently Curated Proteins}

To assess real-world robustness, we evaluated different models on an independent test set—\textit{ds\_rcp} (Recently Curated Proteins)—comprising proteins newly added to UniProtKB/Swiss-Prot between 2018 and 2024. This dataset introduces a temporal distribution shift and includes many enzymes with few or no close homologs in the training set, thereby providing a realistic and challenging evaluation scenario (see Supplementary Figure S3). In such settings, both precision and recall naturally decline, as models must contend with novel sequences and incomplete or evolving annotations.

As shown in Fig.~\ref{fig:res_10_folds}(II), nearly all baseline strategies experienced substantial degradation. Alignment-based methods were most severely affected: the mAccuracy of MSA-via-EC dropped from 82.10\% to 43.43\%, and that of MSA-via-RXN fell from 83.80\% to 50.13\%—both decreasing by over 36 percentage points. Moreover, 4,280 proteins in the test set lacked valid alignments to training sequences, further limiting the applicability of MSA-based approaches (see Supplementary Tables S6–S8). EC-based classifiers also showed pronounced declines. CLEAN, DeepEC, and PRIAM each recorded F1 scores below 36\%, performing worse than the degraded MSA baselines. ECRECer proved relatively more robust, but still suffered ~10 percentage point drops across mAccuracy, mRecall, and mF1, though it remained the strongest EC-based baseline.

By contrast, PLM-based methods exhibited markedly greater stability, particularly in recall (Fig.~\ref{fig:res_10_folds}(II)(b)). All unsupervised PLM variants retained F1 scores above 70\%, following a consistent ranking: UniRep (71.26\%) $<$ ESM (76.08\%) $<$ T5 (77.82\%) $<$ RXNRECer-S1 (80.60\%). A similar trend was observed for mAccuracy, with RXNRECer-S1 again leading at 83.47\%. These results highlight the varying representational capacities of different PLMs and underscore the benefits of supervised fine-tuning. Indeed, RXNRECer-S1 outperformed fine-tuned UniRep and T5, achieving 83.47\% mAccuracy, 89.40\% mPrecision, 81.84\% mRecall, and 80.60\% mF1. Although RXNRECer-S1’s F1 score declined by ~15 percentage points relative to its performance on the controlled test set \textit{ds\_rcv}, it consistently delivered the best balance of precision and recall on \textit{ds\_rcp}. This reduction likely reflects two factors: the growing prevalence of low-homology proteins, and incomplete annotation of recently curated enzymes, where valid reaction links may be incorrectly labeled as false positives.

The observed performance degradation highlights the central challenge of real-world deployment, namely preserving both precision and recall under distribution shifts. To overcome this, we introduce in Stage 2 of RXNRECer (RXNRECer-S2), which adopts a dynamic ensemble strategy to integrates diverse classifiers adaptively, providing a more balanced trade-off between accuracy and efficiency, as detailed in the following section.

\subsection{Ensemble Strategy Evaluation: Effects of Dynamic Integration on Model Performance}

\begin{figure}[htbp!]
\centering
\includegraphics[width=1.0\textwidth]{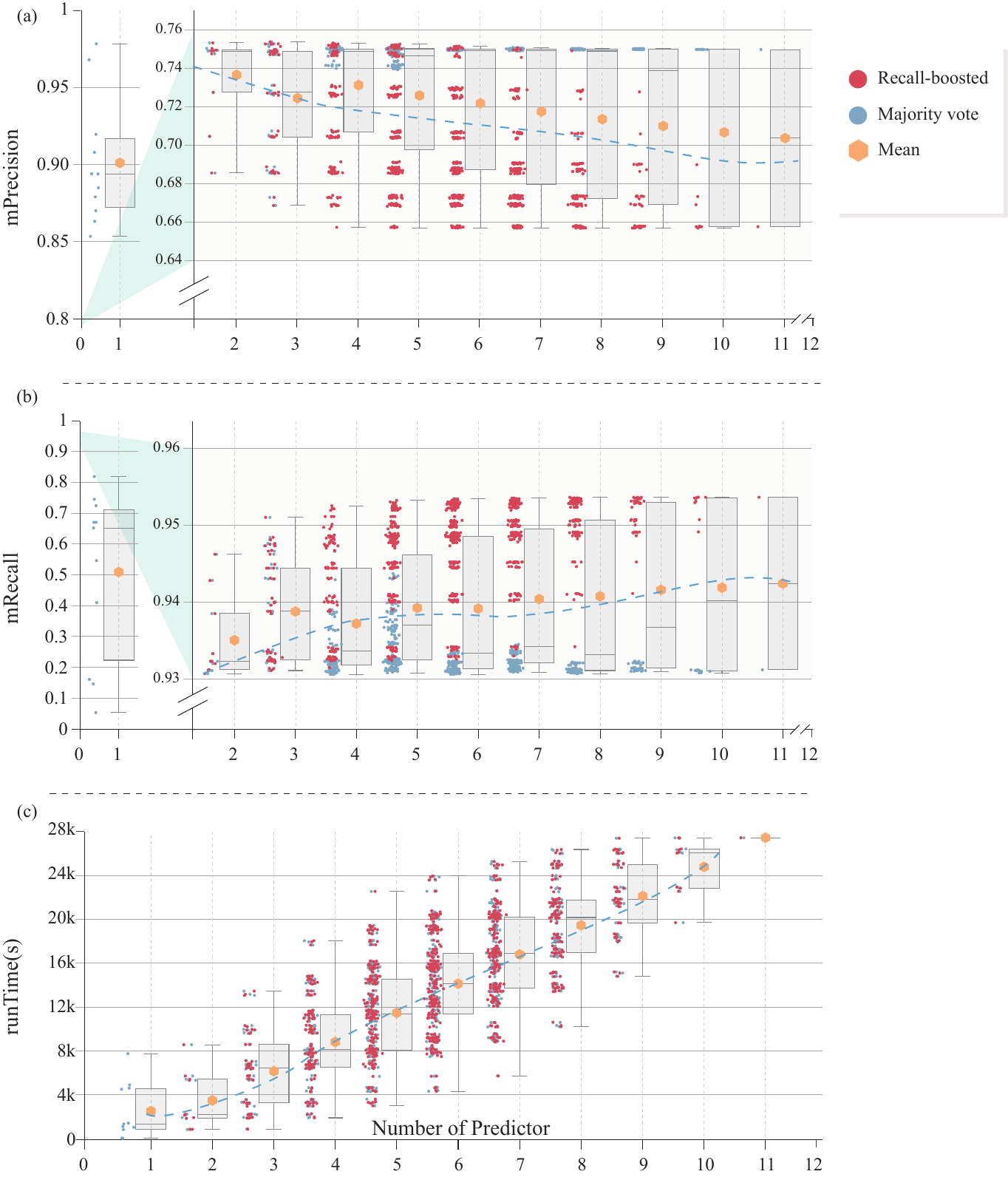}
\caption{
Performance evaluation of ensemble strategies on \textit{ds\_rcp} from three perspectives: 
(a) macro-averaged Precision (mPrecision), 
(b) macro-averaged Recall (mRecall), and 
(c) run time with respect to the number of predictors. 
Boxplots summarize the distribution of results across different ensemble settings, with individual points representing Recall-boosted (red) and Majority-vote (blue) strategies, and orange dots indicating the average values. 
The results highlight the trade-offs between precision, recall, and computational cost when adopting different ensemble strategies.
}
\label{fig:res_ensemble}
\end{figure}

Building on the limitations observed in real-world evaluation, we examined ensemble strategies as a means of balancing performance. Different application scenarios impose distinct requirements on predictive models. In some cases, such as functional annotation pipelines or downstream metabolic modeling, high precision is essential because false positives can propagate and compromise system-level analyses. In other contexts, such as proteome-wide annotation or enzyme discovery in biotechnology, high recall is prioritized to uncover as many potential candidates as possible for subsequent experimental screening, even at the expense of precision. To accommodate these divergent demands, we implements two complementary ensemble strategies: the \textit{Majority-vote} ensemble, which integrates predictions across classifiers to maximize consensus and thereby enhance precision, and the \textit{Recall-boosted} ensemble, which aggregates the union of candidate predictions to maximize coverage.

We systematically evaluated both strategies on the \textit{ds\_rcp} dataset across three dimensions: mPrecision, mRecall, and computational cost. As shown in Fig.~\ref{fig:res_ensemble}(a) and (b), the two strategies reveal clear trade-offs between precision and recall. Ensembles with fewer classifiers tend to achieve higher precision, and under the same ensemble size, Majority-vote consistently outperforms Recall-boosted in precision. By contrast, recall improves as the number of classifiers increases, with Recall-boosted outperforming Majority-vote across nearly all ensemble sizes. These complementary patterns confirm that the two operators serve distinct objectives: Majority-vote favors high-confidence consensus predictions, whereas Recall-boosted favors broad coverage. Importantly, expanding the ensemble with more classifiers does not always yield better results. For example, integrating three methods—RXNRECer-S1, ECRECer, and T5—achieved an mRecall of 94.76\%, whereas a larger ensemble of six methods—RXNRECer-S1, DeepEC, CatFam, PRIAM, ESM, and UniRep—reached only 93.06\%. A similar trend is observed for precision: the three-method ensemble of RXNRECer-S1, ECRECer, and MSA-via-RXN attained an mPrecision of 75.36\%, which is slightly higher than the 75.31\% achieved by the two-method ensemble of RXNRECer-S1 and ECRECer (see Supplementary Tables~S14). These findings highlight that complementarity among classifiers, rather than their sheer number, is the key determinant of ensemble effectiveness.

Runtime analysis further underscores the importance of selective integration (Fig.~\ref{fig:res_ensemble}(c)). Individual predictors span a wide runtime spectrum, which we grouped into five tiers: ultra-fast ($<60$ s; MSAvia-RXN, MSAvia-EC), relatively fast (60–900 s; RXNRECer-S1), fast (900–2000 s; PRIAM, T5-cosine, ESM-cosine, CatFam), moderate (2000–6000 s; DeepEC, ECRECer, UniRep-cosine), and very slow ($>6000$ s; CLEAN, ECPred). For practical large-scale applications, methods exceeding 5000 s are typically excluded due to prohibitive cost (see Supplementary Tables~S13). This analysis confirms that efficiency constraints are as important as predictive gains when designing ensembles.

Considering these three aspects—precision, recall, and efficiency—we developed a dynamic ensemble strategy that adaptively switches between Majority-vote and Recall-boosted integration depending on sequence identity and task requirements. This strategy, implemented as RXNRECer-S2, achieved an mAccuracy of 85.71\%, with mPrecision of 83.44\%, mRecall of 93.06\%, and mF1 of 85.45\% on the \textit{ds\_rcp} dataset, substantially outperforming individual classifiers and fixed-size ensembles. In summary, dynamic integration provides a principled and practical mechanism to reconcile accuracy with efficiency, enabling robust enzyme function prediction across heterogeneous and evolving real-world datasets.

\subsection{Biological Interpretability: A Prompt-Guided Reasoning Approach with Language Models}

While the ensemble strategy in RXNRECer-S2 substantially improved precision and recall, it also expanded the number of candidate reactions per protein, thereby elevating the cost of manual validation. This highlights a practical challenge: beyond quantitative performance, researchers often require biologically interpretable explanations to assess the plausibility of predictions. In real-world annotation tasks, confidence scores alone are rarely sufficient—biologists expect rationales that connect predicted reactions to known sequence features, conserved motifs, or prior biochemical evidence. To address this need, we build RXNRECer Stage 3 (RXNRECer-S3), a prompt-guided reasoning module that leverages large language models (LLMs) not as numerical predictors but as tools to generate structured, human-readable explanations that contextualize predictions in biological terms (Fig.~\ref{fig:framework}(I)(c)). In this way, RXNRECer-S3 complements quantitative predictions with biologically meaningful narratives, thereby reducing manual validation effort and improving the interpretability of reaction-level annotations.

RXNRECer-S3 incorporates a suite of tailored prompt templates (see Supplementary Section~2.4) that guide the language model in interpreting predictions across varying contexts. For proteins with UniProt identifiers (e.g., from TrEMBL; see Supplementary Section~2.2.2), the prompts enable the model to utilize curated annotations, domain signatures, and literature evidence to produce fine-grained justifications. For proteins without UniProt annotations—often from novel experiments or in-house sequencing (see Supplementary Section~2.2.3)—the system applies fallback strategies such as homology search, domain detection, and motif analysis. Each prompt is applied to a structured JSON input containing the protein sequence, UniProt ID (if available), and reaction metadata including RHEA IDs, equations (text, SMILES, ChEBI), and EC numbers. Based on UniProt availability and output type, RXNRECer automatically selects and formats the appropriate prompt. The resulting explanations are concise, biologically grounded summaries that help users evaluate the plausibility of each predicted reaction.

We examined representative examples to evaluate the reasoning capabilities of RXNRECer-S3 across typical and challenging scenarios (see Supplementary Section~2.4.1–2.4.6). The system can reject enzymatic function when catalytic motifs are absent, leverage conserved domains and literature evidence for supported predictions, and generate plausible hypotheses for unannotated proteins using sequence-derived features. It also distinguishes mechanistic subtleties and resolves ambiguous outputs by synthesizing supportive and conflicting evidence. These capabilities enable RXNRECer-S3 to deliver transparent, biologically grounded interpretations that enhance annotation reliability, facilitate experimental validation, and support downstream applications such as pathway modeling and enzyme discovery.

\subsection{Case I: Proteome-Wide Reaction Annotation in \textit{Fusarium venenatum}}

\begin{figure}[htbp]
\centering
\includegraphics[width=1.0\textwidth]{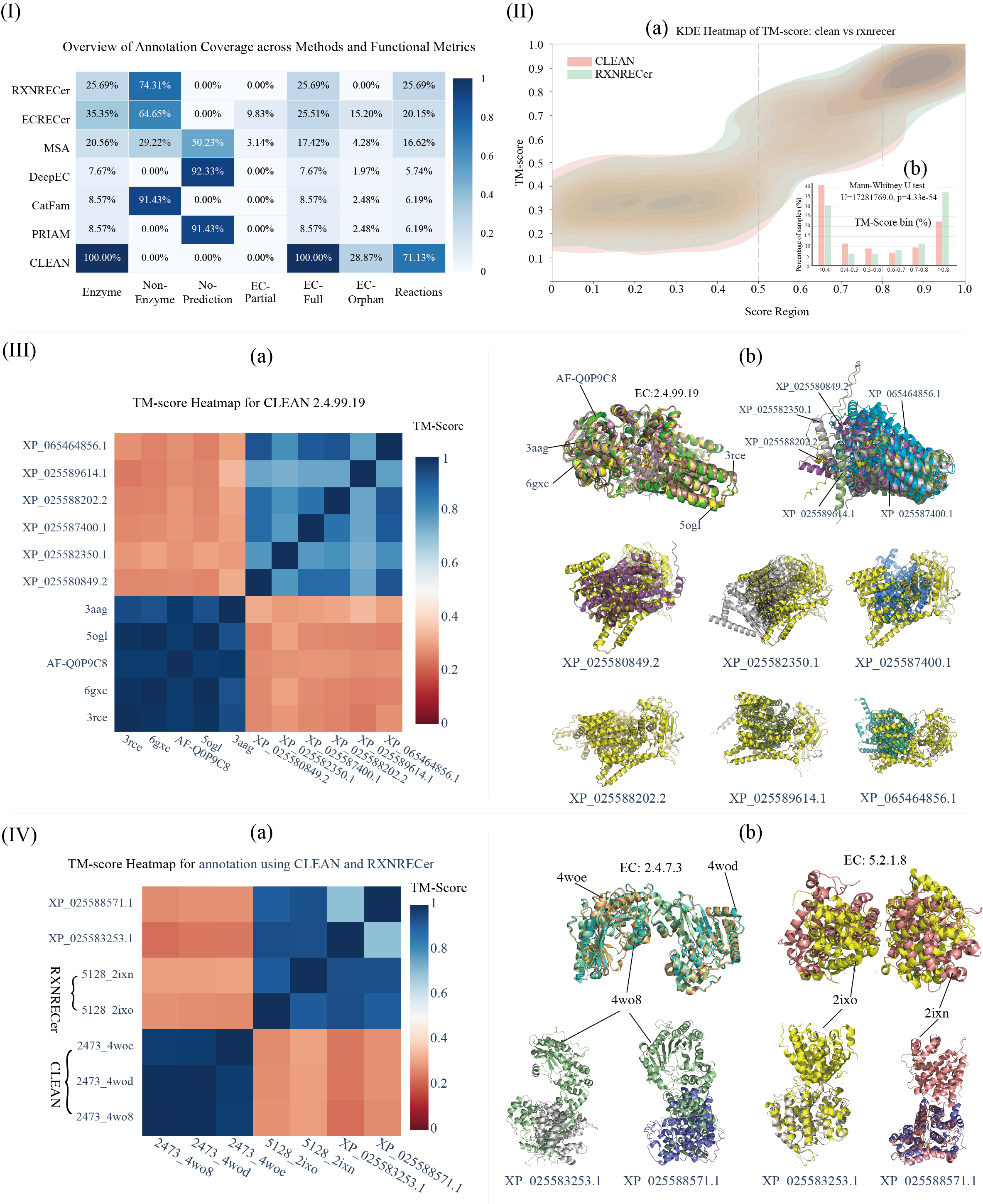}
\caption{
    Case study on large-scale reaction prediction for the \textit{Fusarium venenatum} proteome (FS12832).
    (I) Coverage Comparison. Heatmap showing the proportion of FS12832 proteins assigned to different functional categories by various methods.  
    (II) Structure Consistency Evaluation. 
    (a) Kernel Density Estimation (KDE) plot comparing TM-score distributions between RXNRECer and CLEAN.  
    (b) Bin-wise TM-score comparison and Wilcoxon test showing RXNRECer achieves higher structural consistency.  
    (III) Non-Enzyme Misannotated as Enzyme.  
    (a) TM-score heatmap showing low structural similarity between CLEAN-predicted proteins and canonical enzymes.  
    (b) Structural alignment confirming CLEAN incorrectly assigned EC 2.4.99.19 to non-enzymes.  
    (IV) RXNRECer vs. CLEAN.  
    (a) TM-score heatmap comparing predictions from CLEAN and RXNRECer.  
    (b) Structural alignment showing RXNRECer correctly identifies enzyme functions, while CLEAN fails.
}
\label{fig:case_fusa}
\end{figure}

For newly sequenced organisms, systematic prediction of enzyme-catalyzed reactions is essential to enable metabolic reconstruction and downstream functional studies. \textit{Fusarium venenatum} (FV) strain MPI-CAGE-CH-0201 (GCF\_020744135.1, June 2024) provides a timely example. Members of the \textit{Fusarium} genus are widely recognized for their enzymatic diversity and biotechnological potential, while also attracting attention for their pathogenicity~\cite{dean2012top,pessoa2017fusarium}. Despite this importance, the functional annotation of \textit{Fusarium} proteomes remains sparse, limiting their integration into systems-level analyses~\cite{gnaim2025synthetic,jia2015omics}. To address this gap, we applied RXNRECer-S2 to predict enzyme functions across the FV proteome and compared its outputs with those from established baselines. As shown in Fig.~\ref{fig:case_fusa}(I), the annotation outcomes are summarized into four dimensions: (i) proportions of predicted enzymes and non-enzymes (Enzyme, Non-Enzyme); (ii) frequency of proteins lacking predicted function (No-Prediction); (iii) assignment status of EC numbers (EC-Partial, EC-Full, EC-Orphan); and (iv) proportion of proteins mapped to known biochemical reactions (Reactions).

From Fig.~\ref{fig:case_fusa}(I), we observe that MSA, DeepEC, and PRIAM failed to annotate over 50\% of proteins, limiting their applicability to newly sequenced proteomes. Moreover, all methods except RXNRECer produced orphan or partial ECs that could not be linked to known reactions, thereby constraining their utility for downstream pathway reconstruction. Among methods achieving full proteome coverage, RXNRECer annotated 25.69\% of proteins with known reactions, outperforming ECRECer (20.15\%) while trailing CLEAN (71.13\%). CLEAN’s inflated coverage results from labeling all proteins as enzymes without distinguishing non-enzymes, which comes at the expense of annotation precision.

To assess annotation precision, we examined the structural consistency between predicted enzymes and experimentally validated catalytic templates. For each FS12832 protein, reactions predicted by RXNRECer and CLEAN were mapped to corresponding enzymes in the RCSB PDB. ESMFold was used to generate 3D structures for the query proteins, which were then aligned to their matched templates to compute TM-scores. TM-score is a widely accepted measure of structural similarity, with higher values indicating stronger confidence in functional conservation. TM-scores were categorized into high ($>$0.8), medium (0.5–0.8), and low ($<$0.5) similarity bins ~\cite{zhang2005tm,xu2010significant}. As shown in Fig.~\ref{fig:case_fusa}(II), RXNRECer produced substantially more high-similarity alignments (36\% vs. 22\%) and fewer low-similarity cases (30\% vs. 40\%) compared to CLEAN. Mann–Whitney U test confirmed RXNRECer's significantly higher structural consistency ($U=17281769.0$, $p=4.33 \times 10^{-54}$). Additional comparisons with EC-based methods (ECRECer, CatFam, DeepEC, MSA, and PRIAM) revealed no significant differences, indicating that RXNRECer matches their annotation precision while offering broader reaction-level coverage.

Two representative cases illustrate this advantage. As shown in Fig.~\ref{fig:case_fusa}(III), CLEAN incorrectly assigned several proteins (e.g., XP\_065464856.1, XP\_025589614.1) to lipid-linked N-glycosylation (EC 2.4.99.19), catalyzed by oligosaccharyltransferases.  The FS12832 proteins formed a coherent cluster (TM-scores: 0.72–0.92) but showed low similarity (TM-scores $<$0.35) to canonical PglB enzymes (e.g., 3aag, 5ogl), indicating likely functional misannotation (see Supplementary Table S12). In contrast, Fig.~\ref{fig:case_fusa}(IV) highlights RXNRECer correctly predicting peptidyl-prolyl cis-trans isomerase activity (EC 5.2.1.8, RHEA:16237) for XP\_025588571.1 and XP\_025583253.1. These proteins aligned well with known PPIase structures (e.g., 2ixn, TM-score $>$0.73), whereas CLEAN misassigned them to thiol hydrolases (EC 2.4.7.3) with poor structural similarity (TM-score $<$0.31) (see Supplementary Table S11). These examples highlight RXNRECer’s ability to avoid overprediction and generate biologically plausible annotations.

Overall, RXNRECer achieves a favorable balance between annotation coverage and precision while preserving structural and biochemical consistency. Rather than relying on EC number assignments alone, it directly links predicted enzyme functions to complete and validated biochemical reactions, providing a robust foundation for proteome-scale annotation. These properties position RXNRECer as a powerful tool for metabolic reconstruction, enzyme discovery, and synthetic pathway design in newly sequenced organisms.

\subsection{Case II: From General Reaction Schemas to Substrate-Specific Enzymatic Functions}

\begin{figure}[htbp!]
\centering
\includegraphics[width=1.0\textwidth]{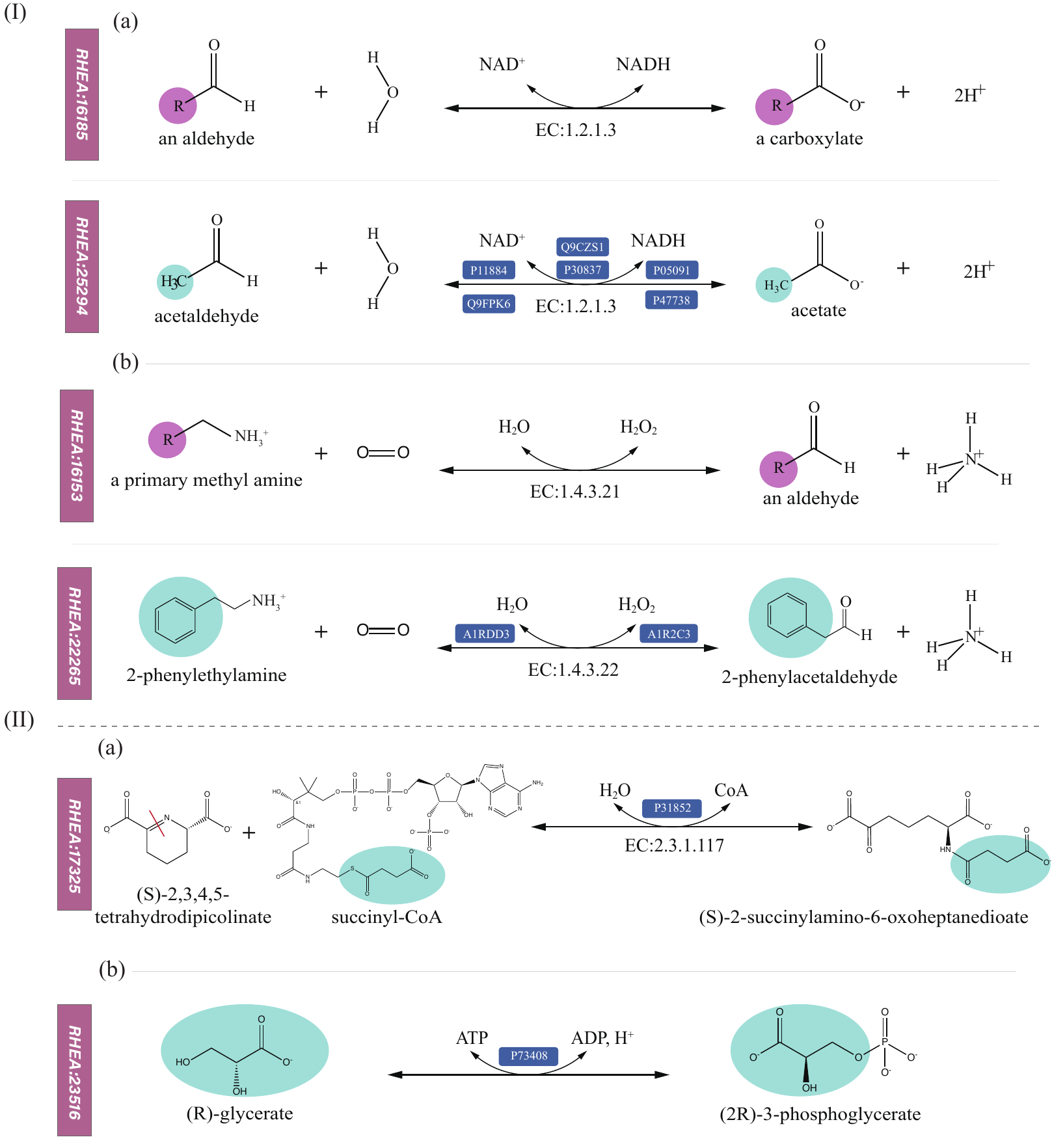}
\caption{
(I) Representative cases illustrating RXNRECer’s capability to resolve general reaction schemas into substrate-specific enzymatic transformations. (a) refines the generic aldehyde oxidation (RHEA:16185) to a specific reaction acting on acetaldehyde (RHEA:25294); (b) resolves a primary amine oxidation template (RHEA:16153) into a specific transformation involving 2-phenylethylamine (RHEA:22265). (II) Representative cases illustrating RXNRECer’s capability to predict reaction-level functions in reviewed proteins lacking curated catalytic annotations, exemplified by P31852 (a) and P73408 (b).
}
\label{fig:case_finegrained}
\end{figure}

Traditional enzyme annotation systems such as the EC classification are designed to organize catalytic functions into hierarchical categories based on overall reaction types. While this schema provides a valuable high-level functional framework, its resolution is often insufficient for downstream biological tasks that require substrate-level specificity. Applications such as metabolic pathway reconstruction, biosynthetic route design, and enzyme engineering frequently rely on precise knowledge of which substrates a given enzyme can act upon. Inaccurate or overly generic annotations may lead to erroneous pathway connectivity, infeasible synthetic routes, or failure to identify viable enzyme candidates for targeted reactions. This limitation is exemplified by the reaction \ce{D-hexose + ATP -> D-hexose-6-phosphate + ADP + H+} (EC:2.7.1.1, RHEA:22740), which broadly describes the phosphorylation of D-hexoses using ATP and encompasses diverse sugars such as glucose, fructose, and mannose. In practice, however, individual enzymes often display strict substrate preferences. Glucokinase Q6CUZ3, for example, catalyzes phosphorylation exclusively on D-glucose or D-mannose and exhibits barely detectable activity toward fructose due to specific substrate recognition and binding constraints \cite{zak2019crystal,kettner2007identification}.

Addressing this resolution gap requires predictive frameworks capable of distinguishing between general reaction templates and precise, substrate-specific activities—a capability that RXNRECer is explicitly designed to offer. As shown in Fig.~\ref{fig:case_finegrained}(I)(a), multiple proteins (e.g., UniProt IDs P30837, P47738, Q9CZS1, Q9FPK6, and P11884; see Supplementary Fig. S7) are associated in UniProt with the general aldehyde oxidation reaction \ce{an aldehyde + NAD^+ + H_2O = a carboxylate + NADH + 2 H^+ } (EC:1.2.1.3, RHEA:16185). RXNRECer not only recovers this broad transformation, but also accurately resolves it to the more specific reaction \ce{acetaldehyde + NAD^+ + H_2O = acetate + NADH + 2 H^+} (RHEA:25294), identifying acetaldehyde as the true substrate. Relevant experimental studies confirm that these enzymes act on acetaldehyde, providing direct support for RXNRECer’s prediction~\cite{larson2007structural, marchitti2007neurotoxicity}. Similarly, as shown in Fig.~\ref{fig:case_finegrained}(I)(b), proteins such as A1RDD3 and A1R2C3 (Supplementary Fig. S3 (II)) are annotated with a generic primary amine oxidation reaction \ce{a primary methyl amine + H_2O + O_2 = an aldehyde + H_2O_2 + NH_4^+} (EC 1.4.3.21, RHEA:16153). RXNRECer further refines this annotation by predicting a specific transformation \ce{2-phenylethylamine + H_2O + O_2 = 2-phenylacetaldehyde + H_2O_2 + NH_4^+}, involving 2-phenylethylamine as the substrate (RHEA:25265), consistent with experimental findings~\cite{lee2013characterization}. These examples highlight RXNRECer’s ability to go beyond general reaction schemes and produce interpretable, biologically grounded predictions at substrate-level resolution.

\subsection{Case III: Predicting Reaction-Level Functions in Proteins Without Curated Catalytic Annotations}

Many proteins in UniProtKB/Swiss-Prot lack assigned enzymatic reactions despite being experimentally verified and structurally characterized. These reviewed entries are often annotated as “uncharacterized” or “with no catalytic activity,” usually due to the absence of curated reaction data or lack of direct biochemical evidence. However, such annotations may overlook latent or mischaracterized enzymatic functions not yet captured in curated databases.

RXNRECer enables the discovery of such hidden functions through direct reaction-level prediction. As shown in Fig.~\ref{fig:case_finegrained}(II)(a), the reviewed protein P31852 (TabB from \textit{Pseudomonas amygdali} pv. \textit{tabaci}) is currently listed in UniProt with minimal functional annotation (annotation score 1/5) and without any assigned reactions. RXNRECer predicts with 99.97\% confidence that TabB catalyzes the transfer of a succinyl group from succinyl-CoA to tetrahydrodipicolinate, yielding N-succinyl-2-amino-6-oxoheptanedioate and CoA (RHEA:17325). This prediction is strongly supported by the biochemical work of Manning et al.~\cite{manning2018functional}, who cloned, expressed, and characterized TabB, demonstrating selective THDPA-\textit{N}-succinyltransferase activity in vitro (see Supplementary Fig.~S8). Sequence and kinetic analyses further confirmed its role in tabtoxin biosynthesis, providing direct experimental validation of RXNRECer’s prediction. Similarly, as shown in Fig.~\ref{fig:case_finegrained}(II)(b), the reviewed protein P73408 (Slr1840 from \textit{Synechocystis} sp.) is also devoid of reaction annotations in UniProt. RXNRECer predicts with 93.76\% confidence that Slr1840 catalyzes the phosphorylation of (R)-glycerate using ATP to produce ADP and 2-phospho-(R)-glycerate (RHEA:23516). This prediction is corroborated by biochemical assays reported by Bartsch et al.~\cite{bartsch2008only}, who demonstrated that recombinant Slr1840 possesses glycerate kinase activity (EC 2.7.1.165). The enzyme’s product specificity was confirmed by coupled assays distinguishing 2-phospho-(R)-glycerate from 3-phosphoglycerate, providing direct evidence for this catalytic function in \textit{Synechocystis} metabolism (see Supplementary Fig.~S9).

These cases illustrate RXNRECer’s capacity to systematically predict reaction-level functions in reviewed proteins lacking curated catalytic annotations, thereby refining enzyme annotation.

\subsection{Case IV: Uncovering Enzyme Promiscuity through Reaction-Level Predictions}

\begin{figure}[htbp!]
\centering
\includegraphics[width=1.0\textwidth]{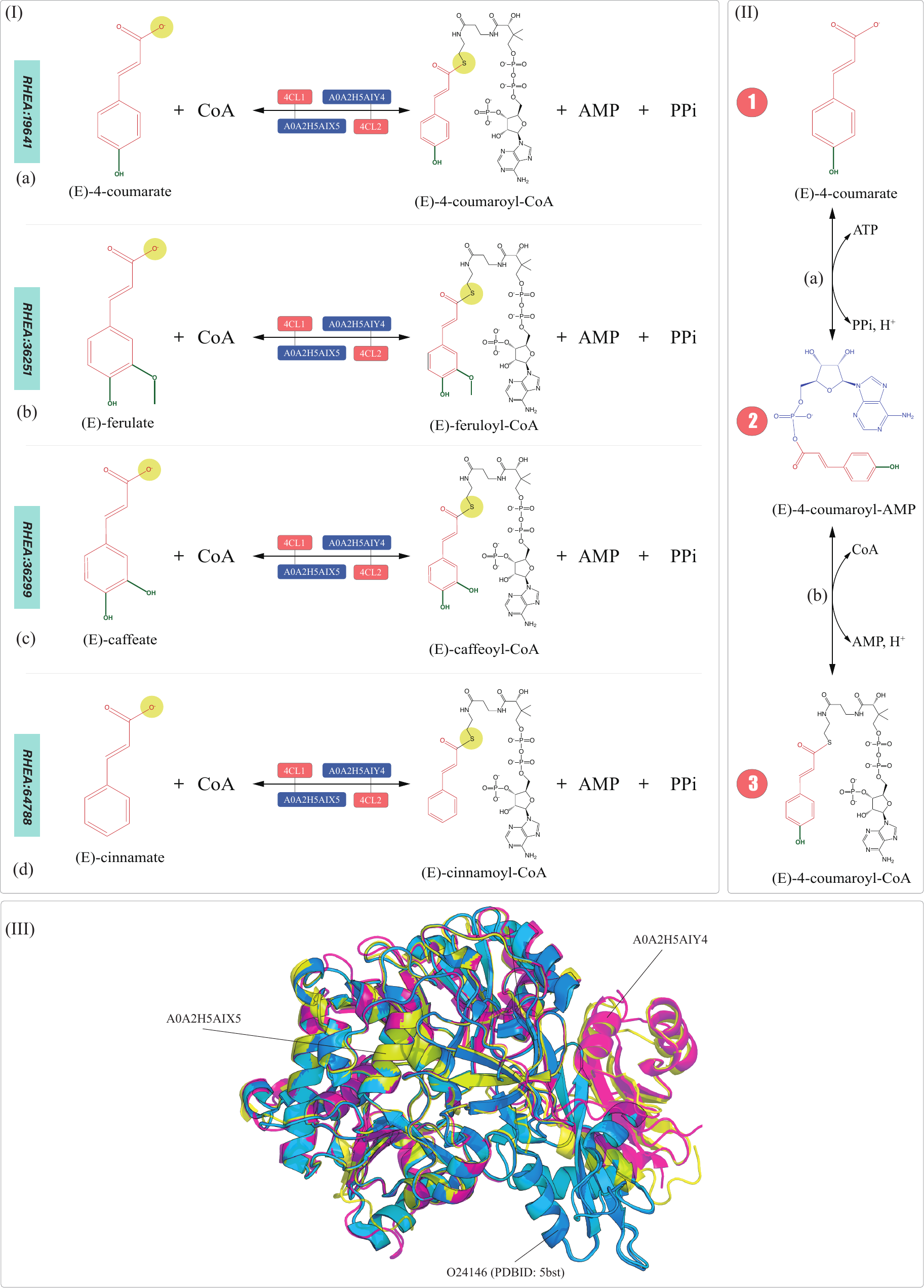}
\caption{
Representative examples illustrating RXNRECer’s ability to uncover enzyme promiscuity through reaction-level predictions. (I) Predicted CoA-ligase reactions catalyzed by two homologous enzymes (A0A2H5AIX5 and A0A2H5AIY4), both originally annotated with a single 4-coumarate: CoA ligase activity (RHEA:19641). RXNRECer further identifies multiple substrate-specific reactions involving (a) 4-coumarate, (b) ferulate, (c) caffeate, and (d) cinnamate. (II) Schematic of the two-step reaction mechanism for (E)-4-coumarate: CoA ligase. The process proceeds via an AMP-linked intermediate: (a) adenylation of the substrate by ATP to form 4-coumaroyl-AMP, and (b) ligation with CoA to yield 4-coumaroyl-CoA. (III) Structural alignment of A0A2H5AIX5 and A0A2H5AIY4 with O24146 (PDB ID: 5bst), a known rice 4CL1 enzyme. The high structural similarity supports the predicted catalytic versatility across different phenylpropanoid substrates.
}
\label{fig:case_promiscuity} 
\end{figure}

Enzyme promiscuity—the ability of a single enzyme to catalyze multiple distinct reactions—represents a valuable yet undercharacterized dimension of protein function~\cite{tawfik2010enzyme}. Recognizing such promiscuous activities expands the known catalytic landscape and informs diverse applications, including metabolic engineering, biocatalyst design, and pathway reconstruction. However, conventional annotation pipelines typically assign only one “canonical” reaction to each enzyme, thereby overlooking alternative functional capacities~\cite{liu2025new}.

RXNRECer directly predicts multiple plausible reaction-level activities for a given protein sequence, providing a powerful means to uncover enzyme promiscuity as a key aspect of functional versatility. As illustrated in Fig.~\ref{fig:case_promiscuity}, two closely related proteins A0A2H5AIX5 (\textit{4CL1}) and A0A2H5AIY4 (\textit{4CL2}) from \textit{Narcissus pseudonarcissus} are annotated in UniProt with a single 4-coumarate: CoA ligase activity (RHEA:19641, EC:6.2.1.12; Fig.~\ref{fig:case_promiscuity}(I)(a)). RXNRECer expands these annotations by predicting a broader spectrum of CoA-ligation reactions with structurally related hydroxycinnamic acid substrates, including ferulate (RHEA:36251), caffeate (RHEA:36299), and cinnamate (RHEA:64788), as illustrated in Fig.~\ref{fig:case_promiscuity}(II--IV)(a).

Alongside predicting multiple reaction-level activities, RXNRECer also resolves the underlying two-step biochemical mechanisms of each predicted transformation. In the first sub-reaction, the enzyme activates the hydroxycinnamic acid substrate using ATP, forming a high-energy acyl-AMP intermediate and releasing pyrophosphate (PP\textsubscript{i}). In the second sub-reaction, CoA acts as a nucleophile and displaces AMP, yielding the final acyl-CoA product. RXNRECer systematically resolves both sub-reactions with mechanistic specificity. For example, the two-step pathway for 4-coumarate corresponds to RHEA:72419 and RHEA:72423; for ferulate, to RHEA:72439 and RHEA:72443; and for caffeate, to RHEA:72431 and RHEA:72435. These steps are illustrated in Fig.~\ref{fig:case_promiscuity}(I–III)(b, c). This mechanistic decomposition reflects the canonical adenylate-forming pathway and demonstrates RXNRECer’s capacity for fine-grained, step-resolved enzyme function prediction.

Biochemical evidence from diverse species indicates that \textit{4CL1} and \textit{4CL2} homologs from multiple species including \textit{Oryza sativa} (rice), \textit{Saccharomyces cerevisiae}, and \textit{Nicotiana tabacum} (tobacco) can catalyze CoA-ligation reactions with structurally related hydroxycinnamic acid substrates, including 4-coumarate, ferulate, caffeate, and cinnamate~\cite{gui2011functional, xin2024engineering, li2015structural}. For example, \textit{Os4CL1} from rice exhibits measurable catalytic efficiencies ($K_\text{cat}/K_m$) toward 4-coumarate ($681~\mathrm{M}^{-1}\mathrm{s}^{-1}$), cinnamate ($633~\mathrm{M}^{-1}\mathrm{s}^{-1}$), ferulate ($824~\mathrm{M}^{-1}\mathrm{s}^{-1}$), and caffeate ($296~\mathrm{M}^{-1}\mathrm{s}^{-1}$)~\cite{xin2024engineering}. Similarly, \textit{Nt4CL2} from tobacco shows activity toward the same substrates, with reported $K_\text{cat}/K_m$ values of $2.65 \times 10^6~\mathrm{M}^{-1}\mathrm{s}^{-1}$ for both 4-coumarate and caffeate, and $1.23 \times 10^6~\mathrm{M}^{-1}\mathrm{s}^{-1}$ for ferulate~\cite{li2015structural}. Consistent with these biochemical findings, structural alignment reveals that the two predicted daffodil enzymes, A0A2H5AIX5 and A0A2H5AIY4, share high structural conservation with a functionally characterized rice 4CL protein (O24146, PDBID: 5bsr), including preservation of key catalytic residues critical for CoA-ligation activity (Fig.~\ref{fig:case_promiscuity}(V)). The concordance between substrate-level biochemical activity across species and structural similarity among homologous enzymes provides compelling support for the functional plausibility of the RXNRECer-predicted CoA-ligation activities.

This case highlights RXNRECer’s ability to uncover latent enzyme promiscuity by jointly predicting multiple substrate-specific activities and resolving their underlying reaction mechanisms.

\section{Discussion and Conclusion}\label{sec_discussion}

In this study, we introduced RXNRECer, a reaction-level prediction framework that offers accurate and interpretable enzyme function annotation by directly linking protein sequences to catalyzed chemical reactions. Unlike traditional EC-level annotation methods, RXNRECer enables more expressive and precise functional assignments by operating at the reaction level. The framework combines PLM-based classification, multi-source ensemble learning, active learning for data-efficient model improvement, and GLM-driven interpretability to deliver accurate, scalable, and biologically grounded predictions across both curated and large-scale proteomes.

Comprehensive evaluations—including 10-fold cross-validation, real-world generalization tests, interpretability assessments, and proteome-wide deployment—demonstrate RXNRECer’s superior performance in terms of accuracy, coverage, and robustness. The dynamic ensemble strategy and active learning loop enhance adaptability to sparse and low-homology scenarios. Meanwhile, the prompt-based reasoning module leverages generalized language models to align predictions with chemically plausible justifications, offering interpretability alongside predictive power.

The case studies further showcase RXNRECer’s practical utility. For example, it successfully annotated thousands of putative enzymes in the Fusarium venenatum proteome, refined ambiguous EC annotations into specific substrate-level reactions, and uncovered latent enzyme promiscuity overlooked by conventional tools. These results validate RXNRECer as a scalable and insightful solution for enzyme function discovery and metabolic network reconstruction.

Nonetheless, certain limitations remain. First, RXNRECer’s predictive scope is still constrained by the diversity and completeness of curated reaction datasets such as Rhea and SwissProt, potentially limiting its ability to capture rare or novel reaction types. Second, although the GLM-based interpretability module provides valuable explanations, its reliance on prompt engineering may introduce variability and lacks grounding in formal mechanistic biochemistry. Third, while the framework performs well across diverse proteins, the absence of explicit structural or ligand-interaction information may limit resolution in cases requiring atomic specificity.

Future directions will focus on expanding the reaction ontology to encompass a broader range of metabolic and synthetic chemistry transformations, incorporating 3D structure-aware modules and biomolecular complex representations to further enhance prediction granularity down to the atomic level. We also plan to explore joint training with metabolic pathway graphs to improve contextual consistency. In parallel, we aim to strengthen the prompt-based reasoning mechanism by grounding it in domain-specific biochemical knowledge, thereby enabling more rigorous and interpretable predictions.

Overall, RXNRECer represents a paradigm shift toward direct, interpretable reaction-level protein annotation. It establishes a strong foundation for advancing enzyme mining, metabolic modeling, and synthetic biology at scale.

\section{Methods}\label{sec_methods}
This section introduces the overall design and underlying principles of the RXNRECer framework. We first formulate the research problem and then describe the proposed method in detail.

\subsection{Problem Formulation}

We formulate the enzyme–reaction prediction task as a multi-label, multi-class classification problem. Let $R = \{r_1, r_2, \cdots, r_n\}$ denote the set of biochemical reactions, and $E = \{e_1, e_2, \cdots, e_m\}$ denote the set of enzymes (proteins). The objective is to learn a function that, given an enzyme $e_j$, predicts its associated reactions from $R$. 

To accommodate both catalytic and non-catalytic proteins, we extend $R$ to include a special virtual reaction “–” representing the absence of catalytic activity. This ensures that all enzymes have at least one assigned label, enabling a unified treatment within a multi-label classification framework. Formally, for each enzyme $e_j \in E$, the model outputs an $n$-dimensional binary vector:
\begin{equation}
  \label{eq:clf_vector}
  clf(e_j) = [y_{1j}, y_{2j}, \cdots, y_{nj}] \in \{0, 1\}^n
\end{equation}
Each dimension $y_{ij}$ of the output vector indicates whether enzyme $e_j$ is predicted to catalyze reaction $r_i$, where $y_{ij} = 1$ if catalysis is predicted, and $y_{ij} = 0$ otherwise.

\subsection{Dataset}

To facilitate a comprehensive evaluation of model performance, we constructed two datasets tailored for distinct purposes: a 10-fold cross-validation dataset (\textit{ds\_rcv}) and an independent temporal test dataset (\textit{ds\_rcp}). The \textit{ds\_rcv} dataset was derived from the UniProtKB/Swiss-Prot snapshot dated January 2018. To ensure data consistency and reliability, we excluded protein entries whose sequences had been deleted or substantially modified in the subsequent January 2024 snapshot. The remaining entries were partitioned into 10 folds using a stratified random split with a 9:1 ratio of training to test instances in each fold, enabling systematic and reproducible benchmarking.  We constructed an independent test set (\textit{ds\_rcp}) from protein entries newly added to UniProtKB/Swiss-Prot between 2018 and 2024. To avoid data leakage, we excluded those overlapping with \textit{ds\_rcv} or associated with reactions already present in its training folds. A detailed description of dataset construction and summary statistics is provided in Supplementary Section~S1.3.

\subsection{PLM-based Reaction Classification}

As illustrated in Fig.~\ref{fig:framework}(I)(a), RXNRECer-S1 is a PLM-based classifier designed to directly predict enzymatic reactions from protein sequences. It leverages a pretrained protein language model, ESM2~\cite{lin2023evolutionary}, to extract rich sequence embeddings that capture amino acid context, global semantics, and co-evolutionary signals. To enable reaction-level multi-label classification, the protein embeddings are passed through a lightweight neural head consisting of a bidirectional GRU, a Transformer layer, and a two-layer feedforward classifier. The model outputs a 10,479-dimensional reaction score vector for each protein. To reduce overfitting and computational cost, we freeze the first 23 layers of ESM2 and fine-tune only the last 10 layers.

To mitigate class imbalance inherent in multi-label tasks, we optimize the model using a variant of focal loss~\cite{lin2017focal}, which down-weights easy negatives and emphasizes hard positive examples. The loss for each label is defined as:
\begin{equation}
\mathcal{L}_{\text{focal}} = -\frac{1}{n} \sum_{i=1}^{n} \alpha_i (1 - \hat{y}_i)^\gamma y_i \log \hat{y}_i + (1 - \alpha_i) \hat{y}_i^\gamma (1 - y_i) \log (1 - \hat{y}_i),
\end{equation}
where $y_i \in \{0,1\}$ is the ground-truth label, $\hat{y}_i \in (0,1)$ is the predicted probability, $\gamma > 0$ is the focusing parameter, and $\alpha_i$ is a class-balancing coefficient derived from label frequency. This helps adjust for label skew across rare and frequent reactions.

This classification module constitutes the backbone of RXNRECer, enabling high-throughput and fine-grained reaction prediction with approximately 208 million trainable parameters (see Supplementary Section~S3.1). To evaluate the impact of partial fine-tuning, we tested multiple layer-freezing strategies and found that unfreezing the final 10 layers consistently yielded the optimal trade-off across standard multi-label metrics (accuracy, precision, recall, F1) (see Supplementary Table~S1).

\subsection{Training using Active Learning Strategy}

Training large-scale models such as RXNRECer is computationally intensive. For instance, the 700M-parameter ESM-2 model required 512 NVIDIA V100 GPUs and eight days to complete 500K updates~\cite{lin2023evolutionary}. By extrapolation, training a 200M-parameter RXNRECer model for a single epoch (500K updates) on an 8×A800 GPU cluster would take over 100 days. To reduce training cost while maximizing learning efficiency, we adopt a multi-strategy active learning (AL) framework that iteratively selects the most informative samples to guide model refinement with minimal supervision.

As illustrated in Fig.~\ref{fig:framework}(III), the training process proceeds through $R$ iterative cycles, each comprising three steps: (i) training on the current set $\mathcal{D}_l$; (ii) selecting informative samples from the untrained pool $\mathcal{D}_u$ using a fine-tuned selector $S(e)$; and (iii) selecting a validation subset from the residual pool via a validation selector $V(e)$ to monitor generalization performance. The process continues until the training budget is exhausted or the model converges. We initialize the training by randomly sampling 1\% of entries from $\mathcal{D}_u$ to construct the initial training set $\mathcal{D}_{L0}$ and train the initial model $M_0$. In subsequent cycles, we sequentially apply two complementary strategies to refine the sample selection process:

\vspace{0.5em}
\textbf{Attention-based Active Sampling.}
This strategy leverages prediction divergence with vs.\ without the attention module. Specifically, for each enzyme $e_i \in \mathcal{D}_u$, we obtain classification outputs from the RXNRECer-S1 framework with and without the transformer-based attention module $A(\cdot)$:

\begin{equation}
\label{eq:active_maxsep_p1}
p_{ei}^{\text{w/o}} = FC(GRU(ESM(e_i)))
\end{equation}

\begin{equation}
\label{eq:active_maxsep_p2}
p_{ei}^{\text{w/}} = FC(A(GRU(ESM(e_i))))
\end{equation}

\begin{equation}
\label{eq:active_maxsep}
\delta_{ei} = \left| p_{ei}^{\text{w/}} - p_{ei}^{\text{w/o}} \right|
\end{equation}

The top-$K$ samples with the highest discrepancy scores $\delta_{ei}$ are selected as the most informative candidates for retraining. This approach effectively prioritizes ambiguous or uncertain examples whose predictions rely heavily on contextual attention signals. The detailed algorithm is provided in Supplementary Material Section~S3, Algorithm~2.

\vspace{0.5em}
\textbf{Clustering-based Informative Sampling.}
To ensure diversity and to focus on poorly performing subspaces, the untrained pool $\mathcal{D}_u$ is partitioned into $m$ clusters ${C_1, C_2, \dots, C_m}$. For each cluster $C_j$, we estimate the average model error and compute a normalized sampling weight:

\begin{equation}
\label{eq:active_clustering}
\alpha_j = \frac{\text{err}_j}{\sum_k \text{err}_k + \epsilon}
\end{equation}

Each cluster contributes $k_j = \left\lfloor K \cdot \alpha_j \right\rfloor$ samples to the selection set, promoting both representativeness and targeted improvement. This strategy complements the attention-based sampling by focusing on distributional diversity and local error signals. After each iteration, selected samples are moved from $\mathcal{D}_u$ to $\mathcal{D}_l$, and the model is retrained. This iterative strategy enables RXNRECer to incrementally enhance performance with minimal training cost, particularly under class imbalance and long-tail conditions. The full procedure is described in Supplementary Material Section~S3, Algorithm~3.

\subsection{Ensemble Prediction via Multi-source Integration}

To improve robustness and mitigate model-specific biases, RXNRECer incorporates an ensemble prediction module (RXNRECer-S2) that integrates outputs from diverse EC-based annotation tools and PLM similarity-based approaches. The EC-based tools include CatFam~\cite{yu2009genome}, PRIAM~\cite{claudel2003enzyme}, DeepEC~\cite{ryu2019deep}, CLEAN~\cite{yu2023enzyme}, and ECRECer~\cite{shi2022ecrecer}. In parallel, similarity-based methods—ESM-cosine, T5-cosine~\cite{prostT5}, and UniRep-cosine~\cite{unirep}—compare query proteins to annotated enzyme–reaction pairs using pretrained embeddings to compute reaction similarity scores.

Each method produces a ranked list of candidate reactions with associated confidence values (defaulting to 1.0 when not explicitly provided). These outputs are integrated through a dynamic ensemble strategy that adapts the fusion mode according to the sequence identity (\textit{identity}) between the query protein and known sequences:

\begin{equation}
	\label{eq:ensemble}
	f(\text{modes}) =
	\left\{
		\begin{aligned}
			&\textit{INTE\_{mv}}, && \text{if } \textit{identity} \geq 75\% \\
			&\textit{INTE\_{re}}, && \text{if } \textit{identity} < 75\% \\
		\end{aligned}
	\right.
\end{equation}

Here, $INTE_{mv}$ and $INTE_{re}$ correspond to Majority Voting and Recall-Boost Ensemble modes, respectively, each addressing different application needs:

\begin{itemize}
    \item \textbf{Majority Voting ($INTE_{mv}$):} Prioritizes precision by selecting the reaction most frequently predicted across all base models. If no reaction appears more than once, the prediction from RXNRECer-S1 is used as fallback. This high-confidence consensus scheme is particularly suited for scenarios where minimizing false positives is critical, such as functional annotation pipelines that support downstream metabolic modeling or pathway reconstruction.
    \item \textbf{Recall-Boost Ensemble ($INTE_{re}$):} Prioritizes coverage by taking the union of all candidate predictions across base models. While this increases the risk of false positives, it broadens the candidate reaction space and is advantageous in exploratory applications such as promiscuous enzyme discovery, large-scale proteome annotation, or de novo pathway design, where maximizing recall is more important than strict precision.
\end{itemize}

The cutoff of 75\% sequence identity was selected based on prior studies showing that enzyme function is generally conserved above 70–80\% identity~\cite{rost1999protein,tian2003nucleotide}. In this work, we adopt 75\% as a practical threshold to balance precision- and recall-oriented integration; however, this parameter can be adjusted by users according to their specific application scenarios. Overall, this multi-source ensemble framework leverages the complementary strengths of individual models and enables flexible trade-offs between precision and recall, thereby enhancing the robustness and adaptability of RXNRECer’s prediction capabilities.

\subsection{GLM-based Re-ranking and Interpretability}

Building upon the quantitative predictions from RXNRECer-S1 (PLM-based classifier) and RXNRECer-S2 (ensemble module), the final component of RXNRECer (Fig.~\ref{fig:framework}(I)(c)) incorporates a general-purpose language model (GLM) to enable fine-grained re-ranking and mechanistic interpretability of candidate enzyme–reaction associations.

This module accepts a structured prompt comprising the protein sequence (or UniProt ID), candidate reactions from upstream modules, and relevant biochemical context such as RHEA IDs, compound names, and SMILES representations. The GLM is prompted to identify the most biologically plausible reactions, re-rank them by likelihood, and generate concise natural-language justifications grounded in mechanistic features—such as catalytic motifs, conserved residues, or domain architecture.

Rather than replacing upstream predictions, this module complements them by providing interpretive rationale, filtering false positives, and prioritizing mechanistically supported candidates. This supports fine-grained refinement and expert validation. As detailed in Supplementary Prompts~1–3 and GPT Cases~1–6, outputs follow a standardized JSON format with confidence scores, interpretive rationale, and ranking, enabling consistent and traceable integration into downstream curation workflows.

\subsection{Compared Baselines}

To comprehensively evaluate the performance of RXNRECer, we compared it against several representative baseline methods spanning EC-based annotation tools, sequence alignment techniques, and PLM-based similarity approaches. Specifically, we included five EC-based methods: ECRECer, CLEAN, PRIAM, DeepEC, and CatFam; a traditional multiple sequence alignment (MSA)-based approach; and six PLM-based similarity methods—T5, ESM, and UniRep embeddings combined with either cosine or Euclidean distance metrics (i.e., T5-cosine, T5-euclidean, ESM-cosine, ESM-euclidean, UniRep-cosine, UniRep-euclidean). A detailed description of each method and its implementation is provided in Supplementary Section~S1 (Related Work).

\subsection{Evaluation Metrics}

To comprehensively assess model performance, we adopt four standard macro-averaged classification metrics: accuracy (mACC), precision (mPR), recall (mRecall), and F1 score (mF1). These metrics are defined as:

\begin{equation}
	\label{lb:macc}
	\mathrm{mACC} = \frac{1}{N} \sum_{i=1}^{N} \mathrm{ACC}_i
\end{equation}

\begin{equation}
	\label{lb:mpr}
	\mathrm{mPR} = \frac{1}{N} \sum_{i=1}^{N} \mathrm{PPV}_i
\end{equation}

\begin{equation}
	\label{lb:mrecall}
	\mathrm{mRecall} = \frac{1}{N} \sum_{i=1}^{N} \mathrm{Recall}_i
\end{equation}

\begin{equation}
	\label{lb:mf1}
	\mathrm{mF1} = \frac{2 \times \mathrm{mPR} \times \mathrm{mRecall}}{\mathrm{mPR} + \mathrm{mRecall}}
\end{equation}

Here, $N$ denotes the total number of classes. For each class $i$, $\mathrm{ACC}_i$, $\mathrm{PPV}_i$ (precision), and $\mathrm{Recall}_i$ are computed in a one-vs-rest manner. The macro-averaging strategy ensures that each class contributes equally to the overall metric, regardless of sample frequency—an essential property when evaluating models under class imbalance, as commonly encountered in enzyme function prediction.

\subsection{Software and Code Availability}

All software components were implemented in Python. The RXNRECer framework is built on the PyTorch library, with some baseline methods implemented using scikit-learn. All datasets, trained models, evaluation results, and source code required to reproduce our results are publicly available at: \url{https://github.com/kingstdio/rxnrecer}.

\section{Web Server Implementation}

The RXNRECer web platform is deployed on a local HPC cluster using a modular and scalable architecture. Bioinformatics tools, including DIAMOND and custom Python workflows, are containerized into portable Singularity images, enabling execution through a single Singularity Image File (.sif). Job scheduling is managed by the Slurm workload manager, which efficiently dispatches and scales tasks based on queue load. A lightweight controller oversees task execution and state tracking. All functionalities are exposed through a RESTful API built with FastAPI and documented using Swagger, supporting job submission, monitoring, and result retrieval. The backend also includes a reaction visualization engine (RDKit + Matplotlib) and integrates a local large language model inference module to provide fine-grained, interpretable predictions.

\section{Supplementary Data}

Supplementary data are available online at: \url{https://github.com/kingstdio/RXNRECer/blob/main/document/supplementary.pdf}

\section{Data Availability}

The data underlying this study are available within the article and its online supplementary materials. The source code of RXNRECer, along with the training datasets and prediction results, can be accessed at: \url{https://rxnrecer.biodesign.ac.cn/}.

\section{Author contributions statement}
Z.S. and X.L. designed and implemented the model, conducted the experiments, analyzed the results and wrote the manuscript. J.Z and B.C fine-tuned the model. H.M., D.W, W.W, F.W and Q.Y. reviewed the manuscript. R.W. and H.L. designed the website.

\section{Funding}
This work was supported by the Strategic Priority Research Program of the Chinese Academy of Sciences [XDC0120201], National Natural Science Foundation of China [32201242, 12326611, 61976067].

\backmatter

\bibliographystyle{unsrt}


\end{document}